\documentclass[11pt]{article}
\usepackage[T1]{fontenc}
\usepackage[utf8]{inputenc}
\usepackage{lmodern}
\usepackage[margin=1in]{geometry}
\usepackage{amsmath,amssymb}
\usepackage{booktabs,array,tabularx}
\usepackage{graphicx}
\usepackage[numbers,sort&compress]{natbib}
\usepackage[hidelinks]{hyperref}
\usepackage{microtype}
\usepackage{placeins}
\usepackage{caption}
\usepackage{xurl}
\hypersetup{pdftitle={Routing Beyond Task Categories: A Cost-Aware LAB-Bench Study},pdfauthor={Ahmed Tariq and Leo Guinan},pdfsubject={A LAB-Bench routing pilot and a prospective comparison under a shared inference budget}}

\title{Routing Beyond Task Categories:\\A Cost-Aware LAB-Bench Study}
\author{Ahmed Tariq\\\small Independent researcher \and Leo Guinan\\\small Idea Nexus Ventures}
\date{12 November 2026}

\begin{document}
\maketitle

\begin{abstract}
We tested whether features of biology questions or of a model's first answer can improve accuracy within an inference budget. In a LAB-Bench pilot, we compared a category router with one that adds six structural features. Neither reached the primary 80\% accuracy target on development. At the 50\% and 60\% secondary targets, both chose Haiku for all 120 held-out questions and scored 71/120. We then froze a new comparison before collecting responses on 144 additional questions. Semantic routing scored 86/144. A policy that could defer after reading Haiku's answer scored 87/144. Both stayed within the inference budget of USD~0.678 per 1,000 queries. Against a 95\% Haiku and 5\% GPT mixture, the accuracy gains were 0.66 percentage points for semantic routing and 1.35 points for deferral. The paired 95\% intervals were $[-0.28,2.25]$ and $[-0.14,3.47]$ points. A simple rule that escalated Haiku nonanswers scored 92/144 at USD~0.710 per 1,000 queries, above the budget. A retrospective nonanswer rule with a calibration-selected escalation probability gave an expected 85.63/144. At the learned deferral policy's observed inference cost, post hoc interpolation gave an expected 86.14/144. In a later development check, Astra and Gemini each scored 30/40. The best tested policy in each router family matched that score. The pair covered 33/40 in hindsight. The routers missed the 32/40 criterion, so the proposed 120-question evaluation did not start.
\end{abstract}

\section{Introduction}

When several language models can answer a biology question, choosing a more expensive one does not guarantee a better answer. We test whether information available at the time of routing can help make that choice. A router can use the question before calling a model. A cascade can inspect the first response before deciding whether to call a second model. A retrospective oracle can see which answers were correct.

We study those choices on LitQA2, ProtocolQA and SeqQA, three tasks from LAB-Bench. Our pilot used 60 development questions and 120 held-out questions. We compared a category router with a version that adds six features of the question and its options. At the two secondary targets that both routers met on development, both chose Haiku for every held-out question.

For the follow-up, we used the 180 pilot questions as historical fitting and calibration data. We tested a semantic router that sees question embeddings and a deferral policy that also reads Haiku's response. Both could choose among the same four model configurations. We locked the fitted policies and analysis before collecting one response per model on each of 144 new questions.

On the new questions, we measured how often the policies switched models and how many additional answers they got right under a common inference budget. Both policies scored slightly higher than a mixture selected on the calibration set. The paired intervals include zero. The simpler rule of escalating every Haiku nonanswer recovered more answers and exceeded the inference ceiling.

We then screened three newer models on the original development questions and tested whether routing between the two strongest candidates cleared the accuracy target on the tasks with unused questions.

\begin{table}[htbp]
\centering
\small
\begin{tabularx}{\linewidth}{@{}>{\raggedright\arraybackslash}p{0.21\linewidth}*{2}{>{\raggedright\arraybackslash}X}@{}}
\toprule
Phase & Data & Decision and scope \\
\midrule
Pilot & 60 development and 120 held-out questions & Neither router met the primary target on development. We evaluated the declared secondary policies on the holdout. \\
Prospective follow-up & 120 historical fitting, 60 historical calibration and 144 new evaluation questions & We locked the original 13 policies and analysis before collecting new answers. We added the randomized nonanswer comparison and decision audit after evaluation. \\
Stronger-model screen & Original 60 development questions, with 40 used for the routing check & The tested mixed routers fell short of 32/40 on development. We stopped before selecting the proposed 120 new questions. \\
\bottomrule
\end{tabularx}
\caption{Data and decision points for each study phase.}
\label{tab:phases}
\end{table}

\section{Related work}

RouterBench evaluates routing with recorded model outcomes and inference costs, including an oracle reference~\cite{routerbench}. LLMRouterBench found that models answered different questions correctly and that several tested routers struggled to improve reliably on a simple baseline~\cite{li2026llmrouterbench}. RouteLLM learns selectors from preference data and reports router inference overhead~\cite{routellm}. FrugalGPT learns cascades that can purchase another model response after an earlier attempt~\cite{frugalgpt}. Learning-to-defer research studies when a predictor should hand a decision to another agent~\cite{mozannar2020}. These studies motivate our fixed-model, mixture and cascade controls. Our follow-up uses the observable first response to estimate the gain from another model call.

IRT-Router models relationships between model abilities and query attributes through item response theory~\cite{irtrouter}. COMPASS uses a language model to gather performance and cost evidence from an external corpus, with optional previews of candidate answers~\cite{compass}. Its evaluation distinguishes model-call spending from the router's inference cost. We use fixed statistical predictors to measure what structural, semantic and response features add on the selected LAB-Bench tasks.

Yang et al. study cost-aware protocol routing on LAB-Bench and other reasoning tasks~\cite{protocolrouting}. They hold the solver family fixed while selecting among collaboration protocols. Their study compares learned text-and-metadata selectors with a retrospective oracle and finds remaining gaps on LAB-Bench. We choose among four model configurations and test each policy on questions outside its fitting and selection data.

Tian et al. found that elicited confidence improved calibration on several factual question-answering tasks~\cite{tian2023confidence}. Xiong et al. evaluated confidence calibration and failure prediction separately and found persistent difficulty on tasks requiring professional knowledge~\cite{xiong2024uncertainty}. Our deferral policy estimates the change in correctness from escalation. Its response features include uncertainty language, without treating it as a calibrated probability.

LAB-Bench measures capabilities relevant to biology research~\cite{labbench}. Its authors describe LitQA2 as a retrieval-dependent task. Our no-tool configuration therefore tests model selection under the fixed multiple-choice protocol, with limited evidence about laboratory use. We pin the public dataset revision and the Inspect Evals implementation to preserve the task inputs and scoring procedure~\cite{labbenchdata,inspectlabbench}.

\section{Pilot data and task preparation}
\label{sec:data}

\subsection{Selection and split}

We pinned \texttt{futurehouse/lab-bench} to revision \texttt{5c77cec6} and selected 60 questions from each of three subsets~\cite{labbenchdata}. A deterministic hash ordering fixed the selected IDs and split. Each subset contributes 20 development questions and 40 held-out questions. The 15-question smoke set contains five development questions per subset and adds no independent observations.

The selection keeps recorded LitQA2 and ProtocolQA source groups on one side of the development/held-out split. SeqQA contributes four questions from each of its 15 subtasks. Every SeqQA subtask appears in development, with one or two development questions. The evaluation therefore concerns new questions within represented subtasks. It provides no test of transfer to an unseen biological capability. The manifest audit found no exact full-input duplicate among selected items and no recorded source group crossing the split. These checks cannot identify every conceptual dependency or paraphrase.

We prepared the complete pinned task before selecting IDs. This order matters because the upstream loader and choice shuffle determine the mapping between letters and option text. We used \texttt{inspect\_ai 0.3.263}, \texttt{inspect\_evals 0.20.0}, and seed 7 for both dataset preparation and choice shuffling~\cite{inspectlabbench}. The adapter retained the official reasoning instruction, multiple-choice format and abstention option. ProtocolQA includes its protocol context before the question. We allowed no external retrieval or tools. This restriction matters particularly for LitQA2, whose task design involves evidence retrieval~\cite{labbench}.

\subsection{Information available to the router}

The collector supplies the task category before inference because it selects the benchmark task. Both routers may use that category. CRR may also use the visible question, any protocol context and all answer options. The router input excludes gold-answer and source-reference metadata, the fine SeqQA subtask, model responses and realized usage or cost. Item IDs serve bookkeeping. The \texttt{source\_group} field serves split checks, fold assignment and resampling. Neither field enters the success predictor or expected-cost table.

The evaluation assumes that an application already knows the task category. Routing unlabelled questions would require a category-assignment procedure and its own cost. We do not estimate that cost here.

\subsection{Scoring and denominator}

We extracted answer choices with Inspect AI 0.3.263 and graded them with the LAB-Bench scorer in Inspect Evals 0.20.0. Appendix~\ref{app:followup} reports the answer extraction sensitivity check. A correct answer receives one and every other assigned outcome receives zero. Wrong answers, abstentions, malformed answers and refusals remain in the denominator. We distinguish correct (C), official incorrect-grade (I) and abstained (N) outcomes in the result table. Refusals belong to I. They do not count as transport failures.

The selected matrix contains 240 development cells and 480 held-out cells, with one observed response per model and question. The 60 superseded Haiku development requests remain acquisition overhead outside this matrix (Section~\ref{sec:haiku-recovery}).

\section{Pilot routers and development selection}
\label{sec:method}

Let $c$ denote the known task category and $m$ a model. For $s_{mc}$ successes among $n_{mc}$ training questions, the category router uses a Beta(1,1) posterior mean,
\begin{equation}
 \widehat p_{mc}=\frac{s_{mc}+1}{n_{mc}+2}.
\end{equation}
Every assigned training question contributes to $n_{mc}$. The posterior mean supplies the routing score. The associated Beta uncertainty does not certify calibration or independence between related questions.

CRR uses the same category estimate as a fixed offset in a model-specific logistic predictor,
\begin{equation}
 \widehat p_{\mathrm{CRR}}(m,x)
 =\sigma\!\left(\operatorname{logit}(\widehat p_{mc})+
 \beta_m^{\mathsf T}z(x)\right),
 \label{eq:crr}
\end{equation}
where $\sigma$ is the logistic function and $z(x)$ contains six features. These features describe input length, numeric and sequence structure, option count and option lengths. Appendix~\ref{app:features} gives the exact definitions.

For each training fit, we center each feature using its category mean and scale the centered values by the pooled training root mean square. A constant feature receives scale one. We fit six coefficients per model by minimizing mean binary log loss plus $\tfrac12\lVert\beta_m\rVert_2^2$. The estimator adds no fitted intercept or category interaction. We fixed the features, ridge coefficient and fitting procedure before collection. The procedure makes no routing-model call and uses no embeddings or learned difficulty labels.

\subsection{Out-of-fold predictions and routing rule}

Five source-group folds produce out-of-fold development predictions. Within each category, the fold allocator orders groups by descending size and then by a deterministic hash. It assigns each group to the smallest current fold, breaking ties by fold number. Each fit uses only the other four folds to estimate category success, feature transformations, coefficients and expected costs.

For both routers, the expected inference cost of model $m$ is its training mean within the task category under the frozen token-price schedule. This estimate does not predict question-specific response length. At threshold $t$, the router chooses the cheapest model with predicted success at least $t$. It breaks ties by higher predicted success and then model slot. If no model qualifies, it chooses the highest predicted success, then lower expected cost and then slot. The route record marks this fallback explicitly.

For each policy family, we evaluate the fixed grid $t\in\{0,0.05,\ldots,1\}$. At each accuracy target $q$, we select the threshold with the lowest out-of-fold inference cost among candidates whose observed out-of-fold accuracy reaches $q$. Ties favor higher accuracy and then lower threshold. If no threshold qualifies, that policy remains unavailable. Out-of-fold performance supports this selection. It supplies no unbiased performance estimate after threshold selection.

We set the primary target at $q=0.80$ and the secondary targets at $0.50$, $0.60$, $0.70$ and $0.90$. We fitted the final profiles on all 60 development questions and combined them with the out-of-fold-selected thresholds. We then committed the fit, selected policies, visible-input hashes and held-out routes before collecting held-out outcomes. We made no threshold or feature changes after that lock.

\subsection{CPU selection specification and execution}
\label{sec:cpu-deviation}

The written method called for a development-measured, constant per-query CPU cost for each policy family during threshold selection. The executed selector ranked thresholds using inference cost alone. A constant within one family would preserve that family's threshold ordering, but the implementation did not measure or add such a constant during selection. The final held-out comparison includes measured routing CPU. We retain the executed policies and do not revise them using held-out results.

\section{Pilot collection and cost accounting}
\label{sec:execution}

\subsection{Model configurations}

The model roster comprises Gemini 3.5 Flash-Lite, Haiku 5.5, GPT-5.6 Luna and Sonnet 5.5. All conditions request LOW reasoning and an inclusive 8,192-token output cap. The inclusive cap counts reasoning within output. We omit temperature and sampling seed in every condition. Appendix~\ref{app:config} gives the collection modes and standard prices.

We allowed up to three attempts for transient errors. Every selected held-out request completed on its first attempt. The 480 selected held-out records contain no missing cell or validation failure. The official scorer counted five Haiku refusals and three Sonnet refusals as incorrect.

\subsection{Haiku development data}
\label{sec:haiku-recovery}

The initial 60-question Haiku development batch stalled after returning 37 outcomes. Before receiving any replacement responses, we chose to fit on a complete set of 60 synchronous requests. The replacement set scored 36/60. The partial batch scored 27/37. The different denominators prevent a matched comparison of the two modes. We excluded the partial batch from fitting.

We corrected the development-scope and results-export checks before collecting held-out responses. The correction left the data, folds, fitted models, thresholds and routes unchanged.

\subsection{Scientific cost basis}
\label{sec:cost}

We calculate inference costs by applying the frozen standard prices in Table~\ref{tab:prices} to the recorded token usage. Policy comparisons add recorded local CPU. The retrospective oracle uses inference cost alone.

For a selected model attempt, we price uncached input, cached input, cache writes and inclusive output at the applicable rates. The output count already includes reasoning tokens. We include recorded usage from retries and failures.

The policy comparison adds feature extraction, prediction and selection CPU, along with the selected model's recorded client work for preparation, dispatch, polling and parsing. Separate per-model processes avoid allocating another model's process CPU to the selected request. For Batch calls, the recorder allocates shared client work over the observed batch requests. Local routing measurements use two warm-up repetitions and five measured repetitions. The full record contains 5,400 measurements. The analysis avoids counting diagnostic preparation again when live client CPU already covers it.

We value local process CPU at USD\,$0.000024$ per CPU-second using a fixed reference rate. Memory, network charges and request fees fall outside this boundary. Queue and network waits contribute no process CPU. We report frozen-standard inference plus recorded local CPU as a deployment-cost proxy.

Model-matrix acquisition, fitting and one-time setup remain separate from per-query replay. We declared no amortization horizon. The paper makes no online latency claim from Batch timing or matrix replay.

\section{Pilot evaluation}

\subsection{Policy controls and retrospective oracle}

We report all four fixed models and the development-selected fixed-model baseline. At a target, the latter chooses the cheapest model whose development accuracy meets that target. It stays unavailable when none qualifies. The choice never uses held-out correctness.

The cascade chooses Haiku as the cheap model and GPT as the more accurate model from development data. It requests the second stage only when the first response abstains, is malformed or fails. A valid wrong answer cannot trigger escalation because its wrongness requires the gold label. The second stage receives the original independent task prompt. Replay charges both attempts whenever the gate opens. It does not recreate a sequential live run.

The uniform control selects each model with probability one quarter, independently of the item. We calculate its expected accuracy and cost from the complete observed matrix. Pairwise controls evaluate expected accuracy and inference-only cost for every pair at weights $0,0.05,\ldots,1$, giving 126 expected points. These expectations integrate model-choice randomization and use the same recorded responses. Any choice of an attractive mixture after inspecting held-out outcomes remains retrospective.

For target $q$, the inference-only oracle chooses an item-wise assignment that achieves at least $\lceil120q\rceil$ observed successes at minimum total cost. It assumes free knowledge of correctness. Every assigned item still incurs one call, including items that all models miss. The reference fixed cost is the cheapest individual model whose observed held-out accuracy meets $q$. The oracle opportunity subtracts oracle total cost from this eligible fixed-model total cost. A missing fixed reference leaves that gap unavailable even when the oracle reaches $q$. This construction does not estimate a deployable saving.

\subsection{Paired uncertainty and interpretation}

We use 5,000 paired bootstrap resamples with seed 261026. Within each subset, we resample source groups with replacement and retain all questions in each selected group. The same draw evaluates both policies. We average within each subset, then weight the three subset means equally. The held-out data contain 40 recorded source clusters per subset. We report percentile 95\% intervals and mark zero-width intervals as degenerate.

The interval conditions on the development fit, the observed model responses and the recorded CPU measurements. It includes no uncertainty from fitting on a different development set, sampling another response or repeating the CPU measurements. A zero accuracy difference for identical routes produces a degenerate interval under this resampling scheme. That interval supplies no guarantee about new responses.

The frozen exploratory zero-loss rule requires both policies to meet the target and at least two source clusters per subset. Both intervals must be nondegenerate. The accuracy-difference interval must lie entirely at or above zero, and the saving interval must lie above zero. This check supplies no confirmatory non-inferiority claim. Such a claim would require a justified margin and a design with adequate power. We make no multiplicity-adjusted claims for the secondary targets or controls.

\section{Pilot results}
\label{sec:results}

\subsection{Fixed models differ across the three tasks}

GPT has the highest overall observed accuracy, with 82 correct answers out of 120. Haiku has the lowest standard cost per query. Gemini answers 56 questions correctly and has a higher standard cost than Haiku or GPT. The ordering reflects both the standard token rates and the observed output lengths.

\begin{table}[tb]
\centering
\small
\begin{tabular}{@{}lrrrrr@{}}
\toprule
Model & C & I & N & Accuracy (\%) & USD/1,000 \\
\midrule
Gemini 3.5 Flash-Lite & 56 & 37 & 27 & 46.67 & 2.063998 \\
Haiku 5.5 & 71 & 40 & 9 & 59.17 & 0.627299 \\
GPT-5.6 Luna & 82 & 31 & 7 & 68.33 & 0.790408 \\
Sonnet 5.5 & 76 & 37 & 7 & 63.33 & 8.240916 \\
\bottomrule
\end{tabular}

\caption{Fixed-model held-out outcomes. C counts correct answers, I counts official incorrect-grade outcomes including refusals, and N counts abstentions. Accuracy uses all 120 assigned questions. Cost includes frozen-standard inference and recorded local CPU, in USD per 1,000 queries.}
\label{tab:fixed}
\end{table}

\begin{figure}[tb]
\centering
\includegraphics[width=\linewidth]{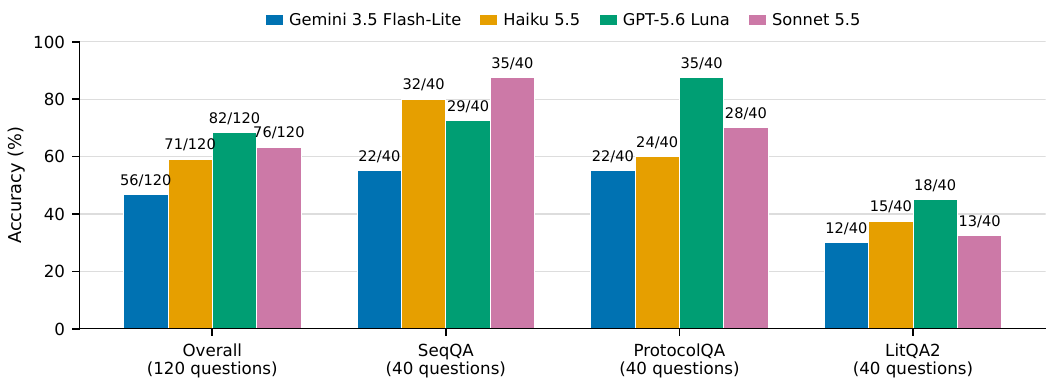}
\caption{Fixed-model accuracy on the complete held-out set and its three 40-question subsets. Labels give correct/assigned counts. Every wrong answer, abstention and refusal remains in the denominator. The plot reports observed accuracies without adding model-level uncertainty intervals.}
\label{fig:accuracy}
\end{figure}

Performance varies across subsets (Figure~\ref{fig:accuracy}). GPT scores 35/40 on ProtocolQA, compared with 24/40 for Haiku. Sonnet scores 35/40 on SeqQA, compared with 29/40 for GPT. All models score below 50\% on LitQA2. These counts show response heterogeneity in this sample. They do not establish that the six visible features can predict which model will answer a particular question correctly.

\subsection{Development accuracy and available policies}

The primary target requires 48 correct answers on the 60 development questions. Across the frozen threshold grid, the category router peaks at 39/60 (65\%) and structural CRR at 42/60 (70\%) on out-of-fold predictions. This leaves the 80\% comparison without a qualifying policy. A retrospective check shows that at least one of the four models answered correctly on exactly 48 development questions. An oracle would have to choose a correct response on every solvable question to reach 80\% on those observed draws.

Table~\ref{tab:targets} includes every declared target. At 50\%, development selects threshold 0.40 for category and 0.00 for CRR. At 60\%, it selects 0.45 and 0.30. All four resulting held-out policies choose Haiku on every question. Each scores 71/120, or 59.17\%. Both 60\% policies fail their held-out target.

\begin{table}[tb]
\centering
\small
\begin{tabular}{@{}lrlrrr@{}}
\toprule
Policy & Target & Available & Correct & Acc. (\%) & USD/1,000 \\
\midrule
Category & 0.50 & Yes & 71/120 & 59.17 & 0.62729921 \\
CRR & 0.50 & Yes & 71/120 & 59.17 & 0.62731007 \\
Category & 0.60 & Yes & 71/120 & 59.17 & 0.62729927 \\
CRR & 0.60 & Yes & 71/120 & 59.17 & 0.62731011 \\
Category & 0.70 & No & n/a & n/a & n/a \\
CRR & 0.70 & Yes & 65/120 & 54.17 & 2.95685522 \\
Category & 0.80 & No & n/a & n/a & n/a \\
CRR & 0.80 & No & n/a & n/a & n/a \\
Category & 0.90 & No & n/a & n/a & n/a \\
CRR & 0.90 & No & n/a & n/a & n/a \\
\bottomrule
\end{tabular}

\caption{Policy availability after development selection and subsequent held-out results. The primary target is 0.80. ``No'' means development supplies no policy. Both routers at 0.50 and 0.60 choose Haiku on every held-out question. Cost uses the same basis as Table~\ref{tab:fixed}.}
\label{tab:targets}
\end{table}

At the 70\% target, development selects a CRR threshold of 0.90. Every held-out question uses the fallback, yielding 65/120 (54.17\%) at USD\,2.956855 per 1,000 queries. The category router has no eligible policy at 70\%. Both policy families remain unavailable at 90\%.

\subsection{Added CPU gives no incremental saving for identical routes}

At the two available paired targets, equal model selections imply equal observed correctness and selected-model inference costs. The only cost difference comes from local routing work. Category costs USD\,0.62729921 per 1,000 queries at the 50\% target, compared with USD\,0.62731007 for CRR. The category-minus-CRR saving is negative, approximately $-1.09\times10^{-8}$ USD per query. The difference is tiny under the declared CPU conversion. It should not carry a practical performance claim.

\begin{table}[tb]
\centering
\small
\begin{tabular}{@{}rrrrr@{}}
\toprule
Target & $\Delta$ accuracy (pp) & 95\% CI (pp) & Saving & 95\% CI \\
 &  &  & \multicolumn{2}{c}{($10^{-8}$ USD/query)} \\
\midrule
0.50 & 0.00 & [0.00, 0.00]$^\dagger$ & -1.0858 & [-1.2980, -0.8863] \\
0.60 & 0.00 & [0.00, 0.00]$^\dagger$ & -1.0836 & [-1.2955, -0.8825] \\
\bottomrule
\end{tabular}

\caption{Paired secondary comparisons. Saving means category cost minus CRR cost, so negative values indicate higher CRR cost. All 5,000 resamples are valid. The dagger marks degenerate accuracy intervals because both policies select the same model on every observed question. These intervals omit repeated model responses and development-fit uncertainty.}
\label{tab:paired}
\end{table}

Both accuracy-difference intervals equal $[0,0]$. The saving intervals lie below zero (Table~\ref{tab:paired}). Neither paired result met the exploratory rule for a cost saving without an accuracy loss. Adding the fitted features left every model choice unchanged at both paired targets. These results provide no test of whether the policies will have equal accuracy on new questions.

\begin{figure}[tb]
\centering
\includegraphics[width=\linewidth]{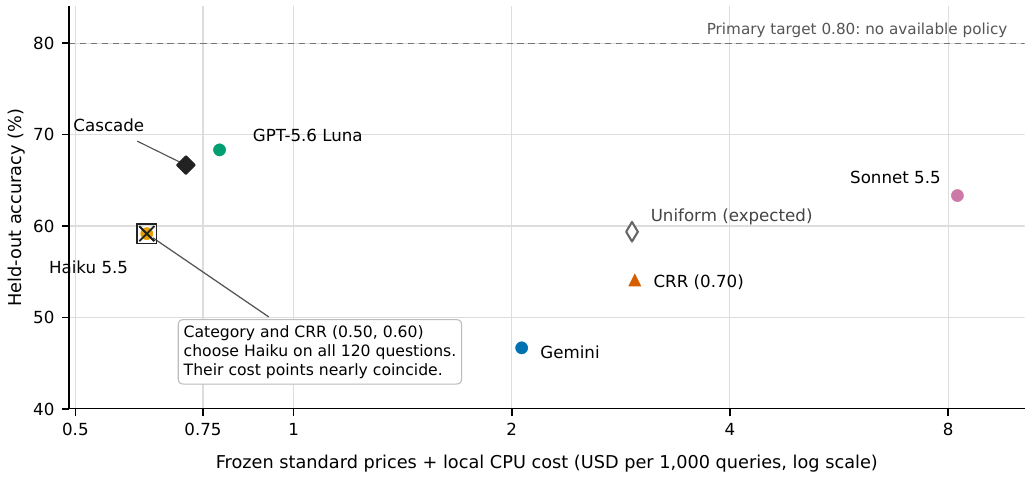}
\caption{Held-out accuracy and frozen-standard inference plus recorded local CPU cost. The horizontal axis uses a logarithmic scale and USD per 1,000 queries. Category and CRR at targets 0.50 and 0.60 nearly coincide with Haiku because all choose Haiku on every question. The plot uses exact coordinates without jitter. Uniform denotes an analytic expectation over model choice. Neither primary 0.80 policy supplies a point.}
\label{fig:cost}
\end{figure}

\subsection{Secondary controls and oracle opportunity}

The development-selected fixed model at 50\% and 60\% is Haiku, which also scores 71/120. No fixed model qualifies on development at 70\% or above. The observable Haiku-to-GPT cascade scores 80/120 (66.67\%) at USD\,0.710357 per 1,000 queries. Uniform selection has expected accuracy 59.375\% and expected cost USD\,2.930655 per 1,000 queries. Figure~\ref{fig:cost} places these controls beside the deterministic policies. Their observed or expected coordinates do not authorize a new held-out-selected policy.

\par\medskip
\noindent\begin{minipage}{\linewidth}
\centering
\small
\captionsetup{hypcap=false}
\begin{tabular}{@{}rrrr@{}}
\toprule
Target & Fixed total (USD) & Oracle total (USD) & Gap total (USD) \\
\midrule
0.50 & 0.07524210 & 0.06529945 & 0.00994265 \\
0.60 & 0.09482385 & 0.06529945 & 0.02952440 \\
0.70 & n/a & 0.06601640 & n/a \\
0.80 & n/a & 0.06948670 & n/a \\
0.90 & n/a & n/a & n/a \\
\bottomrule
\end{tabular}

\captionof{table}{Retrospective inference cost totals over all 120 questions. The oracle uses observed correctness. The oracle reaches both 0.70 and 0.80, where no fixed model qualifies for a cost-gap calculation. Costs are total USD and cover inference alone.}
\label{tab:oracle}
\end{minipage}
\par\medskip

The oracle reaches 80\% at a total inference cost of USD\,0.06948670 (Table~\ref{tab:oracle}). No individual fixed model reaches that target, so the matched oracle opportunity remains unavailable. At 50\% and 60\%, eligible fixed models exist and the respective total oracle gaps are USD\,0.00994265 and USD\,0.02952440. These retrospective gaps use information that the routers do not possess. They do not restore the unavailable primary policy comparison or show that the declared features can recover the gap.

\FloatBarrier
\section{Prospective follow-up on 144 new questions}
\label{sec:followup}

The pilot used structural features of the questions. In the follow-up, we tested whether question embeddings help choose a model and whether features of the first response help decide when to call a second model. Both policies used the same inference budget. We used all 180 pilot questions as historical data and split them by recorded source component into 120 fitting and 60 calibration questions. This pool includes the pilot's held-out questions. We designed the follow-up after exploratory tests on cached outcomes, then locked the fits, policy choices, evaluation cohort and analysis before sending the first new request on 11 November 2026.

\subsection{Cohort and collection}

We used the same LAB-Bench revision, prompt template, option-ordering procedure and scorer as in the pilot. We removed the 180 pilot IDs and candidates connected to those IDs by recorded source references or duplicate full question-and-option content. We chose 48 items from each task. The cohort includes all 48 eligible ProtocolQA items and covers all 15 SeqQA subtasks. Three SeqQA items share a historical question stem with different answer options. ProtocolQA and SeqQA lack underlying source references, which limits the overlap checks.

The four configurations in Appendix~\ref{app:config} each supplied one outcome on every new question, for 576 cells. We retained their collection modes, LOW reasoning and the inclusive 8,192-token output cap. We collected GPT first and then the other three conditions. Appendix~\ref{app:followup} reports the collection outcomes. We retained all five refusals as incorrect outcomes.

\subsection{Policies and budget selection}

Both learned policies estimate how much an alternative model changes correctness relative to Haiku. For each alternative $m$, we fit a ridge regression with penalty 10 to the historical target $Y_m-Y_h$, where $Y_h$ is Haiku's binary correctness. The regressions use one-hot task labels and up to 32 principal components of standardized question embeddings. We compute the embeddings locally with the pinned all-MiniLM-L6-v2 checkpoint~\cite{minilmcheckpoint}. The input contains the full question, any protocol context and the ordered options. We standardize inputs and fit the principal components on the 120 fitting questions.

Semantic routing chooses a single model before generation. It maximizes $\widehat g_m-\lambda(\widehat c_m-\widehat c_h)$, where $\widehat g_m$ is the estimated correctness gain and $\widehat c_m$ is the fitting-set mean inference cost for model $m$ in the task category. Output-aware deferral calls Haiku first. Its predictor also sees 16 features of the visible Haiku response, including response length, answer status, uncertainty markers and whether the parser found an offered choice. It maximizes $\widehat g_m-\lambda\widehat c_m$ over the alternatives, with zero utility for keeping Haiku's answer. A second model receives the original task prompt. The analysis charges both calls when deferral occurs. The routing inputs exclude the gold answer and any alternative model response.

We set the common ceiling to the larger of the Haiku--GPT mean-cost midpoint on the fitting set and Haiku's mean cost on calibration. The midpoint, USD~0.678045 per 1,000 queries, exceeded the calibration Haiku mean of USD~0.66630. The strict-midpoint check used the same ceiling and produced the same policy choices.

We chose each eligible policy's parameters to maximize calibration accuracy within that ceiling. Ties favored lower cost and then the fixed preference in the policy grid. The calibration-selected mixture assigns 95\% probability to Haiku and 5\% to GPT independently of the question. We report its expected accuracy and cost.

The other controls include the four fixed models, a calibration-selected fixed model, the category and structural CRR estimators refit on the same 120 questions, a uniform mixture, a shuffled-label semantic model, and the simple rule that escalates Haiku nonanswers to GPT. Appendix~\ref{app:followup} gives the selected parameters.

We replayed each fixed policy against the same recorded matrix. Inference costs use the pilot's frozen standard prices in Table~\ref{tab:prices}. The budget applies to inference. We also report a proxy that adds measured local CPU for the encoder, response features, prediction and routing at USD~0.000024 per CPU-second. The proxy excludes client dispatch, network charges and one-time setup. We computed paired 95\% percentile intervals for differences from the selected mixture using 2,000 bootstrap draws. Each draw resamples groups within task and retains the questions in each group. Recorded source groups and exact question stems define 142 groups among the 144 questions. These intervals describe the recorded draws with the fitted policies held fixed.

\subsection{Results on the new cohort}

\begin{table}[htbp]
\centering
\small
\begin{tabularx}{\linewidth}{@{}Xrrrr@{}}
\toprule
Policy & \shortstack[r]{Correct\\(out of 144)} & \shortstack[r]{Accuracy\\(\%)} & \shortstack[r]{Standard inference\\(USD/1,000)} & \shortstack[r]{Inference + CPU\\(USD/1,000)} \\
\midrule
\multicolumn{5}{@{}l}{\textit{Fixed configurations}} \\
Gemini 3.5 Flash-Lite & 72 & 50.00 & 2.022599 & 2.022599 \\
Haiku 5.5 & 85 & 59.03 & 0.637203 & 0.637203 \\
GPT-5.6 Luna & 86 & 59.72 & 0.825743 & 0.825743 \\
Sonnet 5.5 & 99 & 68.75 & 7.857681 & 7.857681 \\
\addlinespace[0.35em]
\multicolumn{5}{@{}l}{\textit{Budget-selected baselines}} \\
Selected fixed (Haiku) & 85 & 59.03 & 0.637203 & 0.637203 \\
Task category & 85 & 59.03 & 0.637203 & 0.637204 \\
Structural CRR & 86 & 59.72 & 0.641172 & 0.641183 \\
Pair mixture (expected) & 85.05 & 59.06 & 0.646630 & 0.646630 \\
\addlinespace[0.35em]
\multicolumn{5}{@{}l}{\textit{Unconstrained controls}} \\
Uniform mixture (expected) & 85.50 & 59.38 & 2.835807 & 2.835807 \\
Nonanswer cascade & 92 & 63.89 & 0.709681 & 0.709681 \\
\addlinespace[0.35em]
\multicolumn{5}{@{}l}{\textit{Learned policies and shuffle control}} \\
Semantic gain & 86 & 59.72 & 0.637881 & 0.639618 \\
Output-aware deferral & 87 & 60.42 & 0.648997 & 0.650736 \\
Shuffled semantic gain & 85 & 59.03 & 0.634341 & 0.636078 \\
\bottomrule
\end{tabularx}

\caption{Accuracy and cost for all 13 frozen follow-up policies. Costs are in USD per 1,000 queries. Mixture rows give expected counts. Both budget rules use the USD~0.678045 inference ceiling and give identical routes, accuracy and inference costs.}
\label{tab:followup-policies}
\end{table}

Among the 13 policies locked before collection, output-aware deferral had the highest observed accuracy within the inference ceiling at 87/144 (60.42\%). It gained 1.35 percentage points over the selected mixture, with a paired 95\% interval of $[-0.14,\,3.47]$ points. Semantic routing reached 86/144 (59.72\%) and gained 0.66 points over the mixture, with an interval of $[-0.28,\,2.25]$. Both accuracy intervals include zero. The semantic policy used USD~0.00875 less per 1,000 queries than the mixture at standard inference prices (paired 95\% interval USD~0.00342 to 0.01345). The small cost difference persisted after adding measured CPU (Table~\ref{tab:followup-policies}).

Semantic routing sent four ProtocolQA questions directly to GPT and rescued one error without losing a correct answer. Structural CRR switched three different questions, all in LitQA2, and rescued one error without losing a correct answer. Deferral called GPT after one Haiku abstention and three refusals, all in LitQA2. Two of those four calls rescued an error. The shuffled-label semantic control scored 85/144, matching fixed Haiku.

The simple nonanswer cascade escalated 16 questions, including the four learned deferrals. It recovered seven correct answers over fixed Haiku and reached 92/144 (63.89\%). Its inference cost of USD~0.70968 per 1,000 exceeded the ceiling. Sonnet had the highest fixed-model accuracy at 99/144 (68.75\%) and used USD~7.85768 per 1,000. The learned policies had access to all four configurations and selected only Haiku and GPT on this cohort.

\subsection{Retrospective comparison with randomized escalation}
\label{sec:random-cascade}

After examining the prospective results, we added a comparator that calls Haiku on every question and calls GPT with a fixed probability $p$ when Haiku returns a nonanswer. The rule uses the observed answer status to determine eligibility and uses the same probability for every eligible question. We chose $p$ using historical calibration costs alone. Fixed Haiku cost USD~0.666295 per 1,000 queries on those 60 questions. Escalating every nonanswer cost USD~0.796123. The frozen ceiling of USD~0.678045 gives $p=0.090501$. We apply that probability to the 144 recorded questions and average over the escalation decisions analytically, charging for both calls whenever the policy escalates. Appendix~\ref{app:followup} gives the calculation and the paired checks.

\begin{table}[htbp]
\centering
\small
\begin{tabularx}{\linewidth}{@{}Xrrrr@{}}
\toprule
Operating point & $p$ & Correct /144 & \shortstack[r]{Inference\\(USD/1,000)} & \shortstack[r]{Inference + CPU\\(USD/1,000)} \\
\midrule
Historical calibration costs & 0.0905 & 85.63 & 0.643763 & 0.643763 \\
Evaluation cost matched to learned deferral & 0.1627 & 86.14 & 0.648997 & 0.648997 \\
Evaluation cost set to the ceiling & 0.5635 & 88.94 & 0.678045 & 0.678045 \\
Learned deferral (observed) & & 87 & 0.648997 & 0.650736 \\
\bottomrule
\end{tabularx}
\caption{Retrospective comparisons on the recorded 144 questions. Randomized rows show expected outcomes. The second and third rows use evaluation-cohort costs. The CPU proxy for the randomized rows uses the measured nonanswer check and route selection from the full cascade.}
\label{tab:random-cascade}
\end{table}

The calibration-selected comparator stayed below the inference ceiling in expectation on the evaluation cohort. Learned deferral scored 0.95 percentage points higher (paired 95\% interval $[-0.45,\,2.95]$ points) and cost USD~0.00523 more per 1,000 queries at standard inference prices (paired 95\% interval USD~$[-0.00526,\,0.01826]$). The bootstrap holds $p$ fixed and compares per-question expected outcomes. A deployed policy would incur the costs of its realized second calls.

To compare the policies at equal spending, we used the observed evaluation costs to interpolate between fixed Haiku and the full cascade. The expected count was 86.14 at the learned deferral policy's inference cost and 88.94 at the higher original ceiling. The difference at matched spending was less than one answer. Both operating points depend on the evaluation cohort. Figure~\ref{fig:nonanswer-interpolation} shows the full interpolation.

\FloatBarrier

\section{Exploratory screen with stronger models}
\label{sec:stronger-screen}

We collected one outcome per question for three newer model configurations on the original 60 development questions. We kept the prompts, choice order and scorer. The requests used low reasoning and an inclusive 8,192-token output cap without tools. Two Astra ProtocolQA requests received policy blocks and count as unsolved. Appendix~\ref{app:config} gives the collection settings and standard prices.

\begin{table}[htbp]
\centering
\small
\begin{tabular}{@{}lrrrr@{}}
\toprule
Model & LitQA2 /20 & ProtocolQA /20 & SeqQA /20 & Total /60 \\
\midrule
GPT-5.6 Luna & 11 & 11 & 17 & 39 \\
Gemini 3.8 Flash & 15 & 16 & 15 & 46 \\
Grok 4.6 & 9 & 16 & 12 & 37 \\
GPT-6 Astra & 14 & 18 & 16 & 48 \\
\bottomrule
\end{tabular}
\caption{Fixed-model scores on the same 60 development questions. Luna was the strongest fixed model in the original pilot's development set.}
\label{tab:stronger-models}
\end{table}

Astra reached 48/60 and Gemini reached 46/60. At least one of them answered correctly on 52/60 questions. We chose these two for a routing check on LitQA2 and SeqQA. The two earlier cohorts had already used all 108 ProtocolQA questions in the pinned release. On the 40 development questions in those two tasks, Astra and Gemini each scored 30/40. The two models both answered 27 questions correctly. Astra alone answered three and Gemini alone answered three. Both missed seven. Their hindsight coverage was 33/40. Reaching 32/40 would have required choosing the correct model on at least five of the six disagreements, even with all 27 jointly correct questions retained.

We fitted the category and structural CRR routers from Section~\ref{sec:method} in five source-grouped folds of eight questions, with four questions per task in each validation fold. The original 21 thresholds yielded four distinct category assignments and five distinct structural CRR assignments. A policy had to use both models and score at least 32 correct before new held-out collection. We used recorded token costs at standard rates in the development fit. Two failed Astra attempts lack usage records. Gemini had the lower expected cost in every training fold and task even when we counted only Astra's known costs, so any additional charges on those attempts leave the route choices unchanged.

Both routers peaked at 30/40. Among thresholds that used both models, the cheapest CRR threshold at that score was 0.65. It chose Astra on three SeqQA questions and Gemini on the other 37. The corresponding category threshold was 0.70, with four Astra routes on SeqQA and 36 Gemini routes. Both selected mixed policies chose Gemini on all six disagreements. Their Astra routes fell on questions where the two models agreed in correctness, so none of those switches rescued or lost an answer relative to Gemini. Neither replay chose Astra on either question with an unpriced attempt. Table~\ref{tab:stronger-gate-costs} gives the known costs. These scores use the development pool that informed model selection. No threshold in either tested family qualified. We stopped before selecting or running the proposed 120-question evaluation.

\begin{table}[htbp]
\centering
\small
\begin{tabular}{@{}lrrr@{}}
\toprule
Development policy & Correct /40 & Known USD over 40 & Relative cost \\
\midrule
Fixed Gemini & 30 & 0.12890325 & $1.00\times$ \\
Fixed Astra & 30 & 0.78981000 & $6.13\times$ \\
Best mixed category ($t=0.70$) & 30 & 0.19067825 & $1.48\times$ \\
Best mixed structural CRR ($t=0.65$) & 30 & 0.28599725 & $2.22\times$ \\
\bottomrule
\end{tabular}
\caption{Standard inference costs for the two-task development replay. Fixed Astra's cost omits two failed attempts with unknown charges. Relative costs use known costs and fixed Gemini as the reference.}
\label{tab:stronger-gate-costs}
\end{table}

\FloatBarrier

\section{Discussion and limitations}

In the pilot, CRR reduced out-of-fold probability error for all four configurations (Table~\ref{tab:brier}). That change left the selected routes identical at the two paired secondary targets. The 80\% target sat at the oracle ceiling on development. The stronger models raised that ceiling on the full development set. ProtocolQA accounted for seven of Astra's nine additional correct answers over Luna, and the pinned release supplied no fresh ProtocolQA questions. The two remaining tasks left a smaller margin for gains from switching models.

The follow-up gained one net answer over fixed Haiku with semantic routing and two with learned deferral. The deferral policy chose four Haiku nonanswers and rescued two. The other 12 nonanswers contained five more recoverable cases. A randomized nonanswer rule using a probability set by historical costs also fit the inference ceiling on the evaluation cohort. At the learned policy's observed inference cost, the interpolated rule averaged 0.86 fewer correct answers. At the full ceiling, it averaged 1.94 more and cost more. These retrospective comparisons leave the contribution of the richer response features unresolved.

The stronger-model check evaluated the tested routing families and threshold grid on development. Hindsight coverage of 33/40 shows that the recorded responses contained enough correct answers for a selector with access to correctness. The six disagreements gave the tested features few opportunities to identify a beneficial switch. The check provides no held-out accuracy estimate for the Astra--Gemini routers.

The routing comparisons fit their policies on small sets. The pilot has 20 development questions per task. The follow-up fits on 40 historical questions per task and uses a category mean to estimate each model's inference cost. The common ceiling permits different realized costs within the budget. Our paired intervals condition on the fitted policies and one recorded response per model and question. They leave uncertainty about performance under new fits or repeated draws. An accuracy equivalence claim would need a justified margin and an evaluation sized for it.

The model configurations used different serving setups. We cannot separate their effects from model differences. Identical request settings do not guarantee identical sampling defaults or safety behavior. Some model versions are reported as aliases. Batch timings and request pacing do not measure online latency.

The public questions may overlap model training. Source-group checks cover the references and text we can observe. The pilot and follow-up both cover represented SeqQA subtasks. LitQA2 normally relies on evidence retrieval, which our tool-free condition omits. The results describe multiple-choice performance on this slice. Laboratory practice and clinical use require different tests.

The pilot cost analysis adds recorded local client work and routing CPU. The follow-up adds only the measured CPU for its local features and router. Both use the same reference CPU price. The small cost differences depend on these boundaries.

\section{Conclusion}

The pilot found identical routes at the two paired secondary targets. The follow-up found small gains from semantic routing and output-aware deferral under a common inference budget. Both gains remain uncertain. The full nonanswer cascade recovered seven additional answers over Haiku and exceeded the ceiling. The retrospective randomized comparison leaves open whether the learned policy identifies better escalation opportunities at comparable spending. The stronger-model screen found more correct answers at the model level, while both tested router families fell short of 32/40 on development.

\smallskip
\noindent The \href{https://doi.org/10.5281/zenodo.24220501}{reproduction archive} contains scores, costs and replay code. We used generative AI to help write parts of the collection and analysis code, check artifacts and draft the manuscript.

\FloatBarrier
\pdfbookmark[1]{References}{references}
\bibliographystyle{plainnat}
\begingroup
\small
\setlength{\bibsep}{2pt plus 1pt}
\bibliography{references}
\endgroup

\clearpage
\appendix
\pdfbookmark[0]{Appendices}{appendices}
\section*{Appendices}

\noindent
\hyperref[app:features]{A. Feature definitions}\quad
\hyperref[app:config]{B. Models and prices}\quad
\hyperref[app:subsets]{C. Per-subset results}\quad
\hyperref[app:followup]{D. Follow-up checks}

\section{Exact feature and fitting definitions}
\label{app:features}

Python string lengths count Unicode code points. Let $L$ be input length and let $a_1,\ldots,a_K$ be the lengths of all $K$ answer options, including abstention. The six unnormalized features are:
\begin{align}
 f_1(x)&=\log(1+L),\\
 f_2(x)&=\frac{\#\{\text{input characters satisfying \texttt{str.isdecimal()}}\}}{\max(1,L)},\\
 f_3(x)&=\frac{\text{characters in qualifying sequence tokens}}{\max(1,L)},\\
 f_4(x)&=\log(1+K),\\
 f_5(x)&=\log(1+\overline a),\\
 f_6(x)&=\frac{\max_k a_k-\min_k a_k}{\max(1,\overline a)}.
\end{align}
A qualifying sequence token consists of at least eight ASCII letters from A/C/G/T/U/N, ignoring case, and occupies a whole ASCII-letter token. The sequence proxy measures structure and supplies no established difficulty scale. Option permutation leaves all six features unchanged, while the actual model call retains the official option order.

For feature $j$, the training transform subtracts the category-specific mean $\mu_{cj}$ and divides by a pooled root mean square $s_j$. It sets $s_j=1$ when the feature is constant. Validation and held-out transformations retain those training quantities. Per-model coefficients minimize
\begin{equation}
 -\frac1n\sum_{i=1}^{n}\left[y_i\log p_i+(1-y_i)\log(1-p_i)\right]
 +\frac12\sum_{j=1}^{6}\beta_j^2,
\end{equation}
with the offset in Equation~\ref{eq:crr}. The standard-library solver uses gradient descent with a Hessian-bound step size, maximum absolute gradient tolerance $10^{-9}$ and a 10,000-iteration limit. A convergence failure stops fitting. The artifacts retain coefficients, transformations, fold membership and diagnostics. We fit no additional calibrator and claim no established probability calibration.

Table~\ref{tab:brier} reports the recorded out-of-fold Brier diagnostics. A lower Brier score means lower mean squared error between the predicted probability and binary correctness. The table evaluates development predictions and adds no held-out observations. These values do not establish calibration or identify a new operating point.

\begin{table}[!htb]
\centering
\small
\begin{tabular}{@{}lrr@{}}
\toprule
Model & Category Brier & CRR Brier \\
\midrule
Gemini 3.5 Flash-Lite & 0.263169 & 0.232525 \\
Haiku 5.5 & 0.242593 & 0.224248 \\
GPT-5.6 Luna & 0.230247 & 0.227075 \\
Sonnet 5.5 & 0.263992 & 0.263086 \\
\bottomrule
\end{tabular}

\caption{Recorded development out-of-fold Brier scores for each model's probability predictor. Both methods use the same grouped folds. The table preserves the frozen fit diagnostics and supplies no separate calibration guarantee.}
\label{tab:brier}
\end{table}

\FloatBarrier
\section{Model settings and standard prices}
\label{app:config}

Table~\ref{tab:configuration} lists the pilot configurations. Gemini, Haiku, GPT and Sonnet scored 34, 36, 39 and 37 out of 60 on development, respectively.

\begin{table}[!htb]
\centering
\small
\begin{tabular}{@{}ll@{}}
\toprule
Model & Collection mode \\
\midrule
Gemini 3.5 Flash-Lite & Synchronous \\
Haiku 5.5 & Synchronous \\
GPT-5.6 Luna & Batch \\
Sonnet 5.5 & Batch \\
\bottomrule
\end{tabular}

\caption{Model versions and collection modes in the pilot.}
\label{tab:configuration}
\end{table}

\begin{table}[!htb]
\centering
\small
\begin{tabular}{@{}lrrrr@{}}
\toprule
Model & Input & Output & Cache read & Cache write \\
\midrule
Gemini 3.5 Flash-Lite & 0.3 & 2.5 & 0.03 & n/a \\
Haiku 5.5 & 0.1 & 0.5 & 0.01 & 0.125 \\
GPT-5.6 Luna & 0.2 & 1.2 & 0.02 & 0.25 \\
Sonnet 5.5 & 2 & 10 & 0.1 & 2.5 \\
\bottomrule
\end{tabular}

\caption{Frozen standard prices in USD per million tokens. Inclusive output contains reasoning. The observed Gemini cache-write counts equal zero, so the analysis needs no Gemini cache-write price.}
\label{tab:prices}
\end{table}

The stronger-model screen also used a 20,000-token input guard. We left temperature and sampling seed unset. Gemini and Grok used synchronous requests. Astra used batch requests.

\begin{table}[!htb]
\centering
\small
\begin{tabular}{@{}lrrrr@{}}
\toprule
Configuration and price basis & Input & Inclusive output & Cache read & Cache write \\
\midrule
Gemini 3.8 Flash, standard & 0.75 & 3.75 & 0.075 & 0.041667 \\
Grok 4.6, standard & 2.00 & 6.00 & 0.50 & 2.00 \\
GPT-6 Astra, standard & 10.00 & 50.00 & 1.00 & 12.50 \\
\bottomrule
\end{tabular}
\caption{Frozen stronger-screen prices in USD per million tokens. The two-task gate uses standard prices for both models. The Gemini cache-write rate is rounded here.}
\label{tab:stronger-prices}
\end{table}

\clearpage
\section{Pilot results by task}
\label{app:subsets}

\begin{table}[!htb]
\centering
\small
\begin{tabular}{@{}lrrrr@{}}
\toprule
Model & SeqQA & ProtocolQA & LitQA2 & Overall \\
\midrule
Gemini 3.5 Flash-Lite & 22/40 & 22/40 & 12/40 & 56/120 \\
Haiku 5.5 & 32/40 & 24/40 & 15/40 & 71/120 \\
GPT-5.6 Luna & 29/40 & 35/40 & 18/40 & 82/120 \\
Sonnet 5.5 & 35/40 & 28/40 & 13/40 & 76/120 \\
\bottomrule
\end{tabular}

\caption{Fixed-model correct/assigned counts. Each subset supplies 40 held-out questions. The aggregate denominator includes all 120 questions and gives each subset equal weight.}
\label{tab:subsets}
\end{table}

\FloatBarrier
\section{Follow-up settings and sensitivity checks}
\label{app:followup}

\subsection{Representations and calibration}

The encoder uses the checkpoint in Reference~\cite{minilmcheckpoint}. We tokenize the full question and options without truncation and split the tokens into consecutive windows of at most 254 content tokens. The encoder adds two special tokens per window. We average nonpadding token representations within each window, weight the window vectors by their nonpadding token counts and normalize the resulting 384-dimensional vector to unit length. We run the encoder on CPU in batches of 16 windows. We standardize the embeddings, retain up to 32 principal components and scale the component scores using fitting questions only. Both ridge predictors clip their estimated correctness gains to $[-1,1]$.

The deferral predictor adds 16 observable features. These include log-transformed character, word and choice counts, empty text, answer status, uncertainty, abstention or refusal language, an answer marker, an in-range parsed choice, and multiple distinct option markers. We standardize the response features on the fitting questions. Route-score ties favor the lower incremental cost, then Haiku, then the fixed model-slot order.

The calibration grid contains zero, 81 log-spaced penalties from $10^{-2}$ to $10^{6}$, and a policy that always uses Haiku. The selected semantic penalty is 19,952.623 and the selected deferral penalty is 1,584.893. The category and structural CRR thresholds are 0.35 and 0.30. The selected fixed model is Haiku. The mixture grid covers all six model pairs in steps of 0.05, including the endpoints. The shuffled control permutes all four fitting labels jointly within each task before training the semantic predictors and uses the same calibration rule. It selects a penalty of 10,000. We use seed 20261111 for the deterministic group allocation, label permutation and follow-up bootstrap.

\subsection{Model outcomes and collection}

The replay includes 576 final outcomes. Five Gemini requests and one Haiku request failed before generation and succeeded on retry. We observed no token usage on those six rejected attempts and no truncated responses.

\begin{table}[!htb]
\centering
\small
\begin{tabular}{@{}lrrrr@{}}
\toprule
Fixed configuration & LitQA2 & ProtocolQA & SeqQA & Total \\
\midrule
Gemini 3.5 Flash-Lite & 19/48 & 24/48 & 29/48 & 72/144 \\
Haiku 5.5 & 19/48 & 28/48 & 38/48 & 85/144 \\
GPT-5.6 Luna & 26/48 & 29/48 & 31/48 & 86/144 \\
Sonnet 5.5 & 27/48 & 33/48 & 39/48 & 99/144 \\
\bottomrule
\end{tabular}

\caption{Fixed-model correct counts on the follow-up. Each task has 48 assigned questions. The total uses all 144, including refusals and abstentions.}
\label{tab:followup-subsets}
\end{table}

\subsection{Retrospective replay and sensitivity}

Let $H$ be the inference cost of fixed Haiku per 1,000 queries and $F$ the cost of escalating every Haiku nonanswer. For the historical calibration set we choose $p=\operatorname{clip}_{[0,1]}((B-H)/(F-H))$ at the original ceiling $B$, provided fixed Haiku fits and the incremental cost is positive. A baseline already above $B$ is infeasible. A zero-increment comparison requires no division. For each evaluation question the expected increment is $p$ times the observed difference in correctness and $p$ times the GPT inference cost when Haiku has a nonanswer. The replay includes every question and uses the original task and source-component groups for paired resampling. It prices the local nonanswer check with the recorded full-cascade CPU measurement. It has no separate measurement of the random draw.

\begin{table}[!htb]
\centering
\small
\begin{tabularx}{\linewidth}{@{}Xrrrrrr@{}}
\toprule
Policy & Changed & Rescues & Harms & Both right & Both wrong & \shortstack[r]{Haiku\\nonanswers} \\
\midrule
Category & 0 & 0 & 0 & 0 & 0 & 0 \\
Structural CRR & 3 & 1 & 0 & 0 & 2 & 0 \\
Semantic gain & 4 & 1 & 0 & 1 & 2 & 0 \\
Output-aware deferral & 4 & 2 & 0 & 0 & 2 & 4 \\
Shuffled semantic gain & 16 & 1 & 1 & 5 & 9 & 3 \\
Full nonanswer cascade & 16 & 7 & 0 & 0 & 9 & 16 \\
\bottomrule
\end{tabularx}
\caption{Paired correctness among changed decisions relative to fixed Haiku on the 144 questions. A rescue changes an incorrect result to correct. A harm changes a correct result to incorrect. The reviewer files include task-level counts and costs.}
\label{tab:paired-decisions}
\end{table}

The learned deferral policy selected the four highest estimated GPT gains among the 16 Haiku nonanswers. Those four estimates range from 0.632 to 0.850. GPT answered two correctly. The highest-ranked case that the policy skipped was a Haiku refusal that GPT answered correctly. The estimates describe changes in correctness and take values in $[-1,1]$.

We also replayed the frozen routes after dropping the three SeqQA questions that share historical stems with different answer options. We gave equal weight to the three task means over 48 LitQA2, 48 ProtocolQA and 45 SeqQA questions. Accuracy was 58.56\% for fixed Haiku, 58.62\% in expectation for the selected mixture, 59.26\% for semantic routing and structural CRR, 59.95\% for learned deferral and 63.47\% for the full cascade. The direction of these comparisons stayed the same. The primary follow-up still uses all 144 questions.

\begin{figure}[!htb]
\centering
\includegraphics[width=0.85\linewidth]{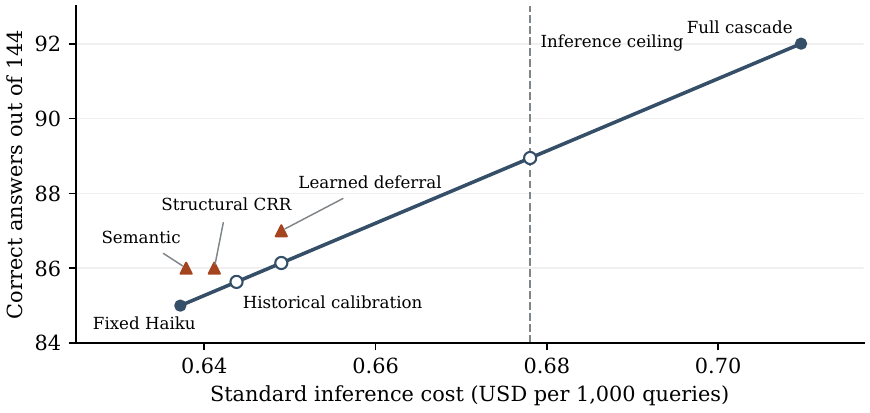}
\caption{Cost and accuracy for randomized escalation on the 144 recorded questions. The line gives expectations from zero to full escalation. The three open circles mark the calibration-selected probability and the two probabilities computed from evaluation costs. The triangles mark observed frozen policies.}
\label{fig:nonanswer-interpolation}
\end{figure}

\FloatBarrier
\paragraph{Answer extraction sensitivity.}
We checked all 1,476 selected outcomes across the pilot, follow-up and stronger-model screen. We replayed 1,460 saved completions with the pinned parser and with the answer-wrapper fix from Inspect AI 0.3.264~\cite{inspectwrapperfix}. Both passes used the original LAB-Bench scorer and prepared choice orders. Sixteen service failures, refusals or blocks bypassed answer extraction. The fix changed zero extracted choices, grades or response classifications. The 16 Haiku nonanswers in the follow-up comprised 12 explicit abstentions and four refusals.

\FloatBarrier

\end{document}